\documentclass[letterpaper, 10pt, conference]{ieeeconf}
\IEEEoverridecommandlockouts
\usepackage{multirow}
\usepackage{lipsum}
\usepackage{amsmath}
\usepackage{cite}
\usepackage{amssymb}
\usepackage[ruled, vlined]{algorithm2e}
\usepackage{graphicx}
\graphicspath{{./figures/}}
\usepackage{booktabs}

\title{\LARGE \bf A Closed-loop, State-centric, Multi-agent Framework for Passenger Load Estimation from Heterogeneous Data Streams}

\author{
\parbox{\textwidth}{%
		\centering
		Yiyao Xu, Hao Zhou, Yuhang Wang, Jingran Sun\textsuperscript{*}%
	}%
      \thanks{%
    All authors are with the Department of Civil and Environmental Engineering, 
    University of South Florida, Tampa, FL 33620, USA
    (e-mail: \{yiyaoxu,haozhou1,yuhangw,jingransun\}@usf.edu).
    \newline\indent
    \textit{*\,Corresponding author: Jingran Sun. jingransun@usf.edu}}%
 }

\begin{document}
	\maketitle
	\thispagestyle{empty}
	\pagestyle{empty}

	\begin{abstract}
		To support operations and passenger-facing services, transit agencies need reliable passenger load trajectories. Currently, load estimates are typically inferred from imperfect sensing systems rather than fully observed, and the accuracy of modern automatic passenger counting (APC) systems still varies with station layout, flow intensity, and operating conditions. To address the challenges of robust passenger load estimation from heterogeneous data streams, including incremental count errors, evidence conflicts, and context-dependent sensor reliability, we propose a closed-loop, state-centric, multi-agent framework. This method enforces physical feasibility at every step, allocates trust dynamically among evidence sources, and feeds physics-derived violation residuals back into training for robustness improvement. The architecture consists of a unified stop-event backbone, a coupled Perception--Physical--Fusion loop for stop-by-stop inference, and optional trip-level macro-correction and closed-loop calibration modules.
	\end{abstract}

\section{Introduction}
\label{sec:introduction}

Public transit agencies need reliable passenger load trajectories for crowding management, dispatching, real-time passenger information, schedule design, and disruption response \cite{Jenelius2020CrowdingInfo, Wang2021TwoStageBPL, Barabino2025APC_Brescia, ref_Optimal_Bus_Bridging_2024, ref_Stochastic_Path_Disruptions_2025}. However, passenger load is rarely observed directly in practice and must be reconstructed from imperfect, heterogeneous sensing streams.

Automatic passenger counting (APC) systems are widely deployed, yet field evaluations show that performance can degrade under crowding, occlusions, and varying station layouts, and systematic biases may appear in operational use \cite{Pronello2023VideoAPC, Barabino2025APC_Brescia}. Learning-based APC methods improve counting quality from sensing signals, but they do not remove the need for careful validation under operating conditions \cite{Seidel2021NAPC, Jahn2022NeuralAPC}. To reduce cost and increase coverage, agencies also consider wireless sensing such as Wi-Fi and Bluetooth probe signals \cite{Demetrio2024WiFiBluetoothToF}. Probe-based counts are not direct passenger counts: they depend on device behavior, randomization, and association uncertainty, and can vary with operating context \cite{He2022Cappuccino, Demetrio2024WiFiBluetoothToF}. Agencies therefore rely on evaluation protocols and recommendations, often together with manual auditing, to assess data quality and support calibration \cite{Boyle2008PassengerCountingSystems, VDV2018Recommendation457v21}.

A key technical difficulty is that passenger load is a latent state governed by stop-by-stop recursion. Commercial APC systems primarily report stop-level boardings and alightings, which behave as increments of a latent load state rather than the state itself. Small per-stop errors can accumulate over successive stops and yield integration drift and physically impossible trajectories \cite{Pronello2023VideoAPC}, such as negative loads or loads above vehicle capacity. These challenges mean that passenger load inference is not only an accuracy problem; it is also a feasibility problem and a data-integrity problem for downstream applications. Open-loop approaches that either (i) predict flows and then cumulatively sum them, or (ii) regress load directly without enforcing recursion constraints, are therefore drift-prone. State-space formulations are also used for efficient online estimation in related transit settings \cite{ref_RealTime_Transit_SSM, ref_SSM_Exponential_2023}, but robustness still depends on how sensing errors and feasibility are handled inside the recursion.

At the same time, the available evidence is heterogeneous. Manual counts are accurate but sparse; Wi-Fi device counts are abundant but indirect; and operating context can shift both demand and sensor error \cite{Pronello2023VideoAPC, Demetrio2024WiFiBluetoothToF}. Given this heterogeneity, a fixed fusion method for all conditions creates a practical failure mode: when an external anchor is reliable, fixed fusion may not suppress recursive drift; when the anchor is missing or anomalous, fixed fusion can inject errors into the state recursion. This motivates a method that (i) enforces physical feasibility throughout state updates and (ii) adapts how much each evidence source is trusted under changing operating contexts. Related ideas from credibility-based weighting and adaptive filtering motivate using explicit, auditable trust signals when measurements can be compromised \cite{ref_Noise_Detection_Reputation_IMM, ref_Adaptive_KF_Credibility_2023}.

Existing research addresses parts of this pipeline. Constrained correction and denoising methods improve physical consistency by projecting flow or load estimates onto feasible sets \cite{Cherrier2023APCDenoisingITSC, deOna2014QualitativeAdjustment, Yin2017L1FlowCorrection}. Network-wide approaches combine Automated Fare Collection (AFC) and partial-fleet APC to infer occupancy for trips without APC coverage \cite{Dib2023UnifiedOccupancyITSC}. Predictive load models support real-time crowding information by forecasting stop-level flows and mapping them to onboard load \cite{Wang2021TwoStageBPL, Jenelius2020CrowdingInfo}. However, these lines of work are often not coupled into a single recursive state framework that enforces feasibility inside the stop-by-stop update, handles sensor disagreement through adaptive trust allocation, and feeds violation signals back into learning.

This paper proposes a closed-loop, state-centric, multi-agent framework for passenger load estimation from heterogeneous data streams. Here ``multi-agent'' refers to a modular operator stack acting on a shared state at each stop: a Perception module proposes boarding and alighting, a Physical module enforces conservation, non-negativity, and capacity bounds during the recursion, and a Trust-aware Fusion module allocates trust to external anchors using an auditable policy. We discretize each trip into stop-to-stop directed segments (stop events define segment boundaries) to form a unified backbone that aligns manual counts, Wi-Fi signals, commercial APC, and exogenous context variables. Physics-derived residuals are recycled into training as bounded reweighting signals to regularize drift-prone contexts. Related ideas of embedding explicit constraints and using residual signals for robustness have also been explored in other ITS settings, such as traffic state estimation and safety-aware control \cite{ref_PIDL_Limitations_2023, ref_PERPL_2024, ref_Logic_Informed_RL_2025}. The contributions are threefold: (i) a segment-based, reproducible inference pipeline that keeps the passenger load state physically feasible throughout estimation, not only after the fact; (ii) an adaptive trust-allocation scheme for multi-source streams that makes disagreement handling explicit and explainable; and (iii) a residual-feedback learning strategy that uses constraint-violation signals to stabilize drift-prone contexts under crowding and noisy sensing.

\section{Methodology}
\label{sec:methodology}

\begin{figure}[!t]
\centering
\includegraphics[width=\linewidth,height=0.42\textheight,keepaspectratio]{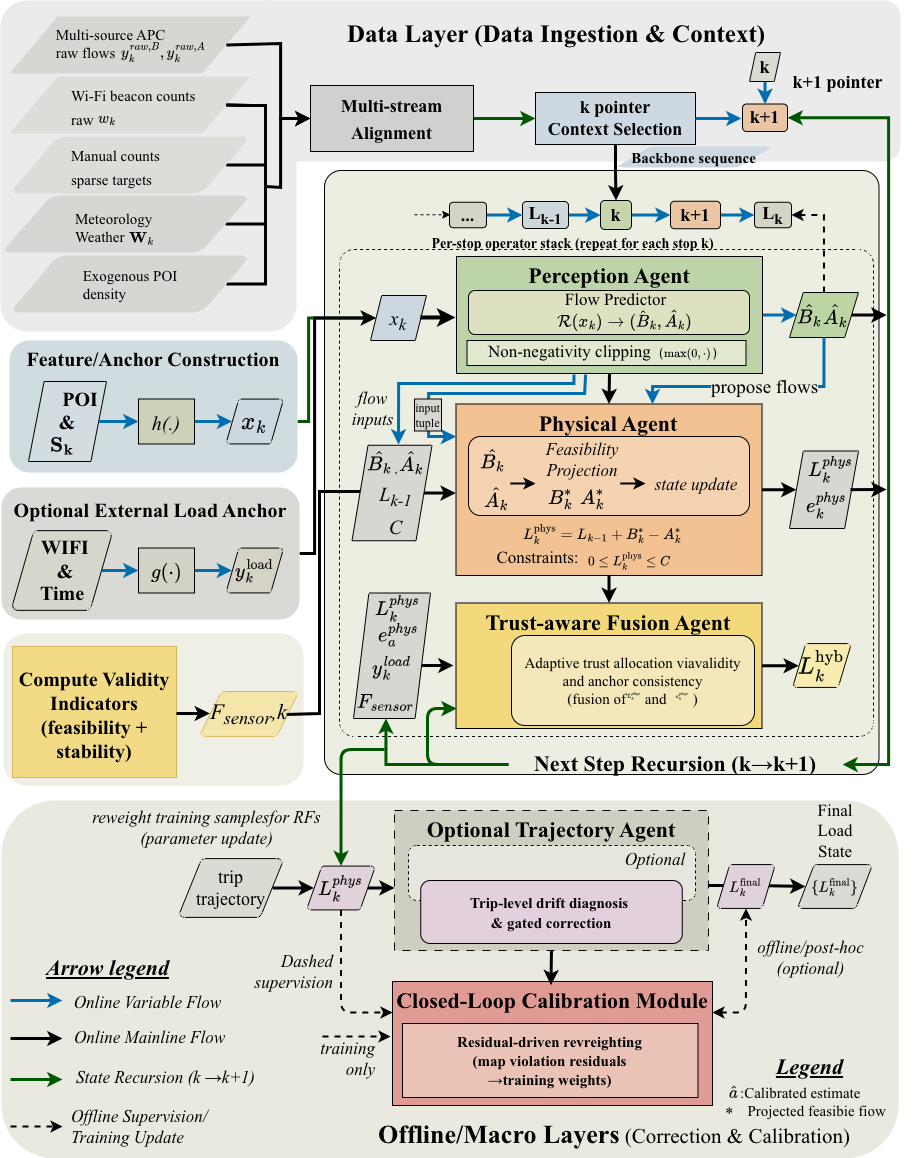}
\caption{State-centric, multi-agent workflow for robust passenger load estimation. Top: data ingestion, multi-stream alignment, stop-indexed context selection, and optional Wi-Fi-derived external anchoring. Center: per-stop agent cascade (Perception $\mathcal{R}$ $\rightarrow$ Physical $\mathcal{P}_C$ $\rightarrow$ Trust-aware Fusion $\mathcal{T}$) repeated for each stop $k$, with the hybrid state $L_k$ written back to drive the recursion $k\!\rightarrow\!k{+}1$; when the external anchor is unavailable, fusion falls back to the physically constrained recursive state. Bottom: optional trip-level macro-correction $\mathcal{M}$ and offline closed-loop calibration via residual-driven reweighting.}
\label{fig:workflow_overview}
\end{figure}

We formulate passenger load estimation as a state-centric recursion over stop events. As shown in Fig.~\ref{fig:workflow_overview}, the framework propagates a latent load state through Perception, Physical, and Trust-aware Fusion agents, with optional macro-correction and residual-driven calibration.

\vspace{-1.0em}

\subsection{Data Alignment and Context Representation}
\label{subsec:data_fusion}
A transit trip is defined as a sequence of stop events $k \in \{0, \dots, K\}$. The latent state $L_k$ denotes the passenger load onboard immediately after servicing stop $k$, initialized at $L_0=0$. We align heterogeneous data streams to this backbone using a temporal tolerance window.

At each stop $k$, the data-driven perception context $x_k$ is formulated as:
\begin{equation}
	x_{k}=\left[\, y_{k}^{raw,B},\; y_{k}^{raw,A},\; \mathcal{T}_{k},\; S_{k},\; \mathcal{W}_{k},\; \mathcal{O}_{occ,k} \,\right]
\end{equation}
where $y_k^{raw}$ are hardware sensor counts, $\mathcal{T}_k$ and $\mathcal{W}_k$ are spatio-temporal and weather features, and $\mathcal{O}_{occ,k}$ is a cross-trip crowding prior. To handle high-dimensional spatial topology, $S_k$ is an interpretable semantic stop label extracted via K-Means clustering on local Point-of-Interest (POI) densities. 

If exogenous Wi-Fi beacon counts $w_k$ are available, they are converted into an external load anchor using an hour-stratified device-to-person mapping:
\begin{equation}
y_k^{\mathrm{load}} = g(w_k,\mathcal{T}_k) = \rho_{h(k)} w_k ,
\end{equation}
where $h(k)$ denotes the hour-of-day bin of stop event $k$, and $\rho_{h(k)}$ is estimated from training trips only and held fixed during evaluation. Anchor availability is encoded by $v_k^{\mathrm{wifi}}\in\{0,1\}$. The resulting anchor $y_k^{\mathrm{load}}$ and validity flag $v_k^{\mathrm{wifi}}$ are excluded from $x_k$ to avoid anchor memorization by the perception model. The clustering for $S_k$ and the anchor mapping $g(\cdot)$ are fitted on training/calibration splits only and held fixed during evaluation.

\subsection{Closed-Loop Micro-Inference Cascade}
\label{subsec:stop_loop}

The core inference engine is a spatial recursion that updates the load state at each stop. We compress the Multi-Agent interactions into a unified nonlinear dynamical system governed by four sequential agents:
\begin{equation}
	\label{eq:stop_loop_cascade}
	\begin{aligned}
	(\hat{B}_{k}, \hat{A}_{k}) &= \mathcal{R}(x_{k}), \\
(L_k^{phys}, e_k^{phys}) &= \mathcal{P}_C\!\left(L_{k-1}, \hat{B}_{k}, \hat{A}_{k}\right), \\
\alpha_k &= \pi\!\left(v_k^{wifi},\, \lvert y_k^{load} - L_k^{phys}\rvert,\, e_k^{phys}\right), \\
L_{k} &= \mathcal{T}\!\left(L_k^{phys}, y_k^{load}, \alpha_k\right).
\end{aligned}
\end{equation}
where the fused state $L_k$ is recursively fed back as the initial condition for step $k+1$. The mechanics of each agent are defined as follows:

\textbf{1) Perception Agent ($\mathcal{R}$):} 
The Perception Agent is a model-agnostic flow predictor that maps the context vector $x_k$ to non-negative boarding and alighting estimates:
\begin{equation}
(\hat{B}_k,\hat{A}_k)=\mathcal{R}(x_k), 
\qquad \hat{B}_k \geq 0,\; \hat{A}_k \geq 0 .
\end{equation}
These data-driven proposals are unconstrained with respect to the recursive load state and may temporarily violate mass conservation. The predictor $\mathcal{R}$ can be implemented using any supervised flow-prediction model.

\textbf{2) Physical Agent ($\mathcal{P}_C$):} Projects the unconstrained proposals onto the physically feasible space defined by vehicle capacity $C$ and non-negativity. Because alighting precedes boarding, the mapping $\mathcal{P}_C$ explicitly clips flows and emits a scalar violation residual $e_k^{phys}$:
\begin{equation}
	\label{eq:physical_projection}
	\mathcal{P}_C: \quad 
	\begin{aligned}
	A_k^* &= \min(\hat{A}_k,\, L_{k-1}) \\[-2pt]
	B_k^* &= \min\!\big(\hat{B}_k,\; C - (L_{k-1} - A_k^*)\big) \\[-2pt]
	L_k^{phys} &= L_{k-1} - A_k^* + B_k^* \\[-2pt]
	e_k^{phys} &= \underbrace{\max(0,\hat{A}_k - A_k^*)}_{\text{over-alighting}} + \underbrace{\max(0,\hat{B}_k - B_k^*)}_{\text{denied-boarding}}
	\end{aligned}
\end{equation}

\textbf{3) Trust Policy ($\pi$) and Fusion ($\mathcal{T}$):} 
The fusion agent $\mathcal{T}$ combines the physically feasible recursive state with the optional external anchor:
\begin{equation}
\begin{aligned}
L_k &= \mathcal{T}\!\left(L_k^{\mathrm{phys}}, y_k^{\mathrm{load}}, \alpha_k\right) \\
&= \mathrm{clamp}_{[0,C]}\!\left(
\alpha_k L_k^{\mathrm{phys}} + (1-\alpha_k)y_k^{\mathrm{load}}
\right).
\end{aligned}
\end{equation}
where $\alpha_k\in[0,1]$ is the physical-state weight. Let
\begin{equation}
d_k = \left|y_k^{\mathrm{load}} - L_k^{\mathrm{phys}}\right|
\end{equation}
denote the disagreement between the external anchor and the physical recursion. The trust policy is defined as
\begin{equation}
\label{eq:trust_alpha_rule}
\begin{aligned}
\omega_k^{\mathrm{wifi}} &=
v_k^{\mathrm{wifi}}
\exp\!\left(-d_k/s_d\right)
\exp\!\left(-e_k^{\mathrm{phys}}/s_e\right), \\
\alpha_k &= \frac{1}{1+\omega_k^{\mathrm{wifi}}}.
\end{aligned}
\end{equation}
where $s_d,s_e>0$ are scale factors. When the anchor is unavailable or strongly inconsistent with the physical recursion, $\omega_k^{\mathrm{wifi}}$ decreases and the update falls back toward $L_k^{\mathrm{phys}}$. When the anchor is valid and consistent, the anchor receives more weight and can correct accumulated drift.
The Physical Agent protects feasibility, while the Fusion Agent controls source reliability by downweighting missing or inconsistent anchors. If all streams are simultaneously biased, the framework can enforce feasibility and provide diagnostics, but it cannot guarantee recovery of the true load without external validation.

\subsection{Macro-Correction and Offline Calibration}
\label{subsec:macro_correction}

\textbf{Trajectory Agent ($\mathcal{M}$):} While $\mathcal{P}_C$ ensures step-wise feasibility, persistent hardware cold-starts can induce parallel trajectory drift. We encapsulate trip-level macro-correction into a gated shift agent $\mathcal{M}$ acting on the entire sequence:
\begin{equation}
\label{eq:macro_operator}
\begin{aligned}
L_k^{final} &= \mathcal{M}\!\left(\mathbf{L}, \hat{\mathbf{B}}, \hat{\mathbf{A}}\right) \\
&= \min\!\left(\max\!\left(L_k + \delta\,\mathbb{I}_{gate},\,0\right),\,C\right).
\end{aligned}
\end{equation}
This binary indicator activates only under persistent, low-variance drift patterns, preventing over-correction of isolated anomalies.

\textbf{Closed-loop Reweighting:} To reduce physically infeasible flow proposals, we recycle inference-time residuals into training. The perception model is refitted using sample weights
\begin{equation}
\omega_k = \min\left(1+\lambda e_k^{\mathrm{phys}}, \omega_{\max}\right),
\end{equation}
where $\lambda$ controls the reweighting strength and $\omega_{\max}$ caps extreme sample weights. This module regularizes the upstream perception model rather than directly correcting the final load state; for example, over-alighting proposals receive larger weights during offline recalibration.

\subsection{Generative Validation Via ABM}
\label{subsec:abm_validation}
To independently audit whether the deterministically reconstructed trajectory $L_k^{final}$ preserves human behavioral randomness, we employ a decoupled Agent-Based Model (ABM). At stop $k$, the ABM generates flows via stochastic priors:
\begin{equation}
	B_{k}^{abm}\sim \mathrm{Poisson}(\lambda_{k}), \quad A_{k}^{abm}\sim \mathrm{Binomial}(L_{k-1}^{abm},\, p_{k})
\end{equation}
The rate parameters $(\lambda_k, p_k)$ are calibrated historically using Bayes-shrunk contextual stratification ($Hour \times S_k$). By evaluating the 1-Wasserstein distance ($W_1$) between $L_k^{\mathrm{final}}$ and the Monte Carlo ABM envelope, we use the ABM layer as a plausibility audit of distributional consistency rather than as a primary predictive component.

\section{Case Study and Experimental Design}
\label{sec:casestudy}

This section evaluates the proposed closed-loop framework on real-world transit data, quantifying how feasibility enforcement, anchoring/trust allocation, and contextual semantics affect cross-trip generalization under strict leakage control.

\subsection{Data Source and Sensing Streams}
The case study uses the operational dataset reported in \cite{Pronello2023VideoAPC}, which evaluated commercial video-based APC errors under real service conditions. Ground truth was collected via onboard manual counting (MC), producing stop-event sequences of boarding, alighting, and onboard load suitable for analyzing integration drift and error accumulation.

The streams serve distinct roles in the pipeline. MC provides supervised targets for Perception training (when used) and the reference load trajectories for evaluation. Video APC supplies raw stop-level flow observations (boardings/alightings) used in the Perception input; vendor-reported load, when available, is treated as a baseline rather than ground truth. Wi-Fi beacons, when available, provide device/MAC counts mapped to external load anchors via a context-calibrated function $g(\cdot)$. Context data (meteorology and POI inventories) provide covariates and spatial descriptors for semantic conditioning. For this study, the capacity bound is set to $C=80$ passengers.

\begin{figure*}[t]
\centering
\includegraphics[width=0.98\textwidth]{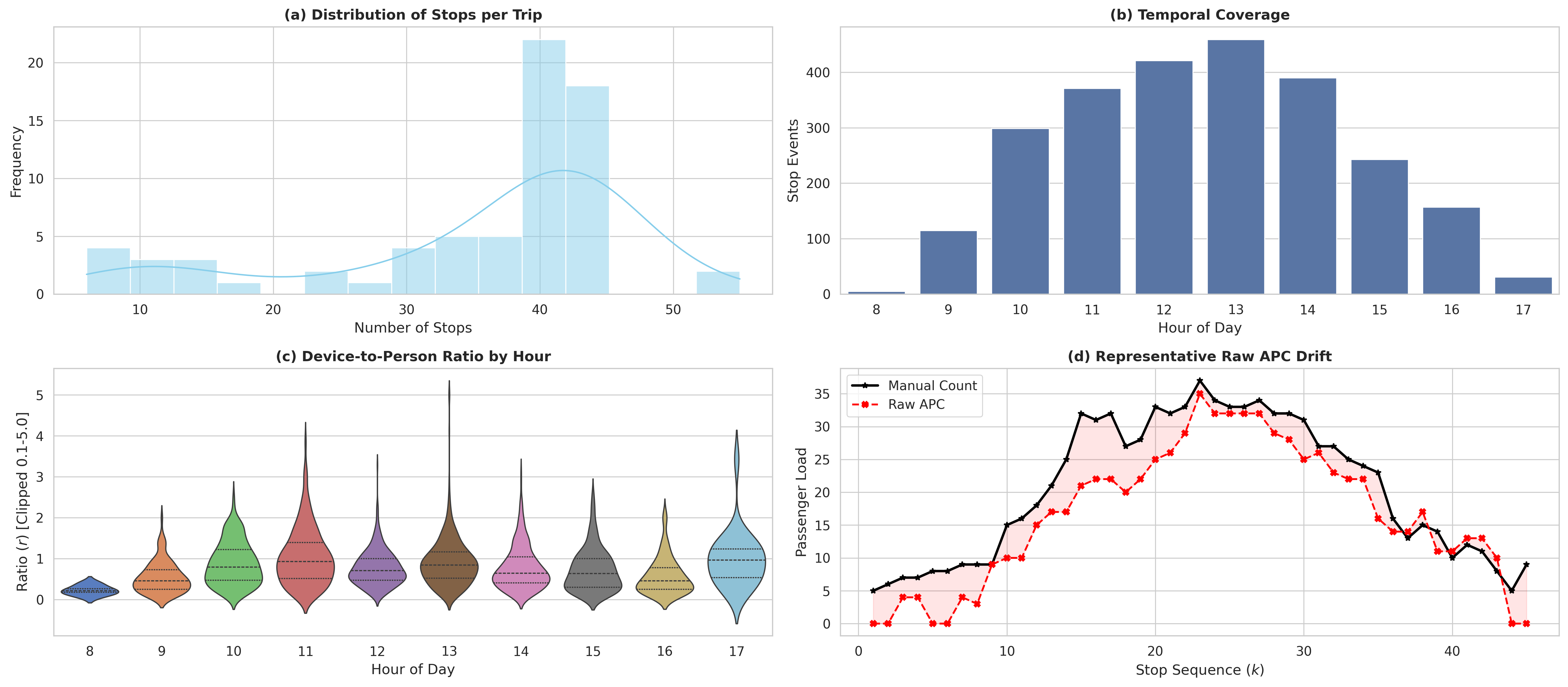}
\caption{Dataset overview and motivating evidence characteristics. (a) Stop count distribution per trip. (b) Temporal coverage across operating hours. (c) Hour-varying device-to-person ratio (clipped to $[0.1,5.0]$), motivating temporal stratification for the Wi-Fi anchor. (d) Representative integration drift of raw APC loads versus manual counts over the stop sequence.}
\label{fig:data_overview}
\end{figure*}

Figure~\ref{fig:data_overview} motivates two design choices: long trips make cumulative drift visible, and the hourly variation in the Wi-Fi device-to-person ratio supports hour-stratified anchor calibration.

\subsection{Repeated Trip-Grouped Cross-Validation and Leakage Control}
To assess generalization and prevent leakage, we adopt a repeated trip-grouped 5-fold cross-validation. A scheduled trip is treated as the atomic unit; all stop events from the same trip are assigned exclusively to either training or test within each split. We repeat the 5-fold split with three random seeds (42, 123, 999), yielding 15 test folds in total.

Within each fold, all train-dependent components are fitted on training trips only and then applied unchanged to test trips. This includes (i) the POI-to-semantics mapping used to assign $S_k$ (when enabled), (ii) the Wi-Fi anchor mapping $g(\cdot)$ calibrated using hour-of-day stratification, and (iii) the Perception regressors trained on MC boarding/alighting targets (when available), with optional residual-driven reweighting computed from a training-only forward pass. All reported metrics are computed on held-out test trips only.

In the present case study, the Perception Agent is implemented with Random Forest regressors because they provide a transparent and low-complexity baseline for testing the closed-loop state-recursion framework under limited operational data.

\subsection{Sensitivity Analysis of Contextual Information}
We ablate contextual inputs to quantify their incremental value, holding constraints, model classes, and gating rules fixed.

\paragraph{Environmental context}
We start from a temporal-plus-crowding baseline and then add meteorological covariates to test whether weather improves robustness under operating conditions that affect both demand and sensing.

\paragraph{Spatial semantics and scale}
We evaluate POI-derived stop semantics $S_k$ learned from POI densities on training trips only. To assess scale sensitivity, we repeat semantic learning using POI densities computed with multiple buffer radii (200\,m, 300\,m, 400\,m); each representation is trained on training trips only and assigned to test trips using the same leakage-control protocol.

\begin{figure}[t]
\centering
\includegraphics[width=0.98\columnwidth]{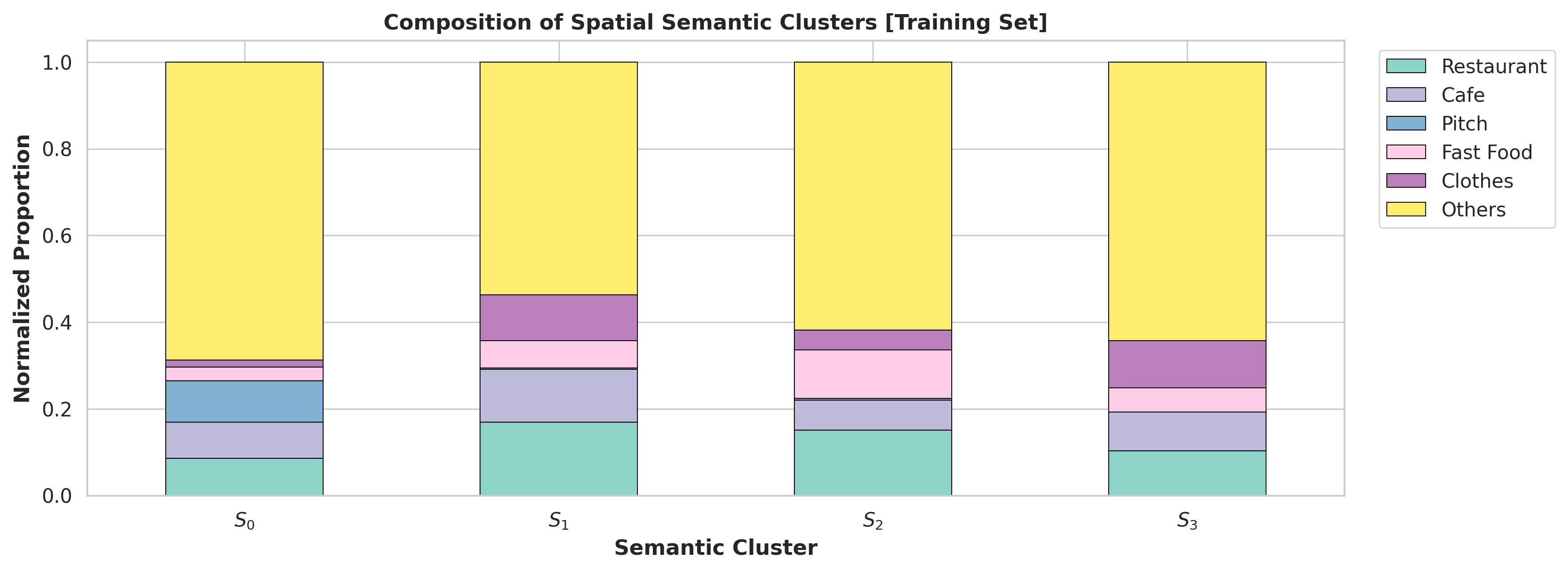}
\caption{Composition of POI-based semantic clusters ($S_k$) learned on training trips only. Distinct cluster profiles support $S_k$ as a compact, interpretable surrogate for high-dimensional spatial context.}
\label{fig:poi_semantics}
\end{figure}

\subsection{Evaluation metrics and reporting protocol}
We report pointwise accuracy between estimated and manual-count load profiles on held-out test trips using RMSE and MAE. To avoid over-weighting long trips, we compute metrics per trip over its stop sequence and then aggregate across trips within a fold; fold-level results are summarized as mean $\pm$ standard deviation across folds. We also report Trip-end absolute error (terminal-stop error) and feasibility/drift diagnostics based on the unconstrained shadow recursion and the physical projection residual rate. All reported metrics are computed on test trips only within each fold and are aggregated across folds as mean $\pm$ standard deviation to summarize cross-trip generalization performance.

\subsection{Representative failure-mode case studies}
Beyond fold-averaged metrics, we analyze representative held-out trips exhibiting distinct failure modes observed in practice: (a) sensor shocks in raw APC, (b) time-local anchor inconsistencies, and (c) distributional regime shifts. To reduce selection bias, we select cases from held-out trips using objective criteria (e.g., high RMSE, high cumulative physics residual $\sum_k e_k^{phys}$, or high frequency of anchor gating), and report the corresponding trajectories together with diagnostic signals that support the interpretation.

\begin{figure*}[t]
\centering
\includegraphics[width=0.98\textwidth]{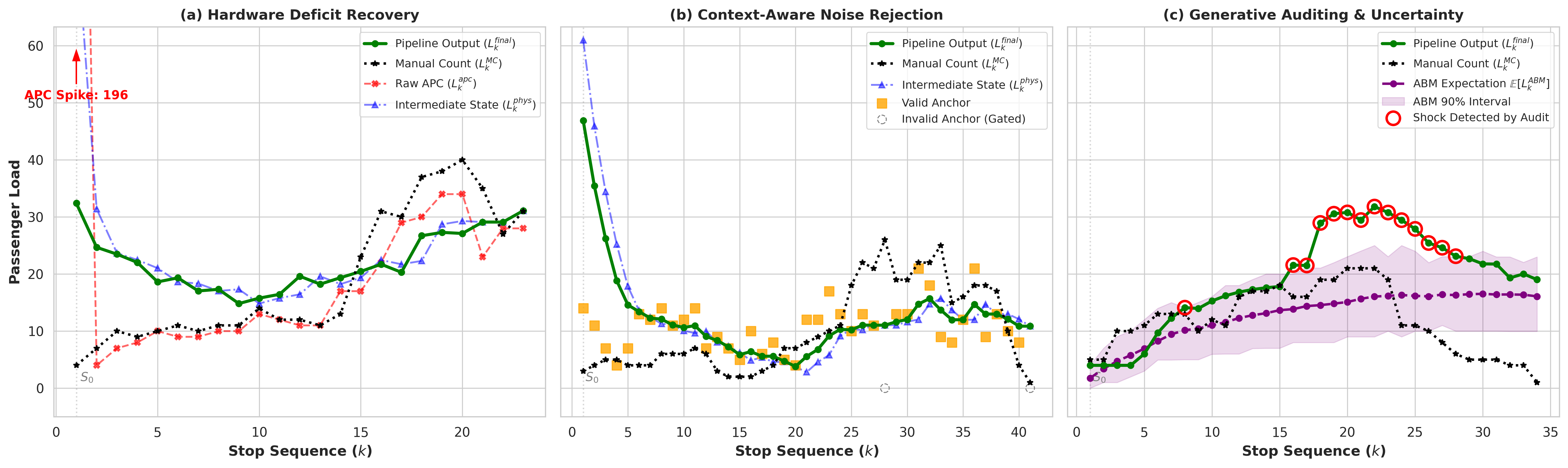}
\caption{Representative failure modes and corrective mechanisms on held-out trips. 
(a) APC spike correction: raw APC (red dashed) induces an implausible transition; the closed-loop output (green) stabilizes the trajectory relative to manual count (black dotted), with the intermediate physics state in blue. 
(b) Context-aware anchor gating: valid anchors (orange squares) are retained while unreliable anchors (hollow markers) are gated; the fused state (green) reconciles anchor and recursive constraints. 
(c) Generative audit: ABM expectation (purple dashed) with 90\% interval (shaded) provides a plausibility envelope; red circles mark detected shocks.}
\label{fig:mechanism_cases}
\end{figure*}

Figure~\ref{fig:mechanism_cases} illustrates three mechanisms. In (a), feasibility enforcement limits the propagation of a large raw-APC shock under recursion. In (b), anchor gating and trust allocation downweight or reject time-local anchor inconsistencies. In (c), the ABM audit provides a context-conditioned plausibility envelope; large deviations trigger an interpretable diagnostic signal for inspection.

\section{Results}
\label{sec:results}
We first describe the evaluation protocol and stress-test setup, and then report accuracy, feasibility diagnostics, and ablation results on both the full test set and APC-inconsistent trips.
\subsection{Evaluation protocol}
We evaluate segment-level passenger load estimates against manual-count load ($mc\_load$) using RMSE and MAE. We also report the absolute error at the final stop of each trip (Trip-end AE). To diagnose integration drift, we compute the infeasibility rate of the \emph{unconstrained} shadow trajectory (Shadow infeas., \%), defined as the percentage of segments where $L^{shadow}\notin[0,C]$, with $L^{shadow}_k=L_0+\sum_{t\le k}(\hat{B}_t-\hat{A}_t)$. In addition, we report the physical projection residual rate ($e_{\mathrm{phys}}$ rate, \%), defined as the percentage of segments with $e_{\mathrm{phys},k}>0$ (over-alighting and/or denied-boarding residual). The capacity is fixed to $C=80$ passengers.

All results use trip-grouped repeated cross-validation with 5 folds and three split seeds (42, 123, 999), totaling 15 test folds. All train-only components (POI semantic clustering, Wi-Fi anchor mapping $g(\cdot)$, the APC-bad threshold) are fitted on training trips and then applied to test trips. We report mean $\pm$ standard deviation across folds (std computed with ddof=1). 

To stress-test the method under poor load consistency, we define an APC-only inconsistency score using the raw APC streams alone. Specifically, we run the same stop-level physical projection on the raw APC boardings/alightings and compute the resulting APC-only physical residuals $e_{k,\mathrm{APC}}^{phys}$. We then define the trip-level APC inconsistency rate as
\begin{equation}
r_{\mathrm{APC}}^{phys}(\mathrm{trip}) :=
\frac{1}{K}\sum_{k=1}^{K}\mathbf{1}\!\left\{e_{k,\mathrm{APC}}^{phys}>0\right\}.
\label{eq:apc_bad_rate}
\end{equation}
A trip is labeled APC-bad if $r_{\mathrm{APC}}^{phys}(\mathrm{trip}) > \tau_{\mathrm{bad}}$, where the threshold $\tau_{\mathrm{bad}}$ is estimated from training trips only and then applied unchanged to the test trips in the same fold.

For the load-inconsistent stress test, only 8 of the 15 folds contain at least three bad trips; stress-test aggregates are computed over these 8 folds.

\subsection{Main results on all test trips}

Table~\ref{tab:main_all_short} summarizes results on all test trips. The proposed stop-level closed-loop system (rule-based trust fusion with physical projection; no trajectory shift) achieves RMSE $9.13 \pm 0.90$ and MAE $7.04 \pm 0.62$. This is substantially better than the perception-only open-loop baseline (RMSE $20.57 \pm 5.29$), confirming that small flow biases can accumulate into large integration drift when load is reconstructed purely by cumulative summation. The phys-only variant without fusion also fails (RMSE $19.98 \pm 5.46$), showing that hard feasibility projection alone is insufficient when evidence streams disagree. These results suggest that the proposed framework can provide a physically consistent and robust load estimate in cases where the load channel is missing, unreliable, or in conflict with other evidence (flows and anchors).

\begin{table*}[t]
\centering
\small
\caption{All trips: mean $\pm$ std across 15 folds (std uses ddof=1). Rates are percentages; the shift rate is trip-level.}
\label{tab:main_all_short}
\resizebox{\textwidth}{!}{%
\setlength{\tabcolsep}{2pt}
\begin{tabular}{lrrrrrr}
\toprule
Method & RMSE & MAE & Trip-end AE & Shadow infeas. & $e_{\mathrm{phys}}$ rate & Shift rate \\
\midrule
Perception-only (open-loop) & 20.57 $\pm$ 5.29 & 13.72 $\pm$ 3.37 & 12.35 $\pm$ 4.82 & 28.0 $\pm$ 7.0 & 0.0 $\pm$ 0.0 & 0.0 $\pm$ 0.0 \\
No fusion (phys-only) & 19.98 $\pm$ 5.46 & 12.71 $\pm$ 3.55 & 13.94 $\pm$ 4.97 & 22.6 $\pm$ 6.4 & 14.6 $\pm$ 4.2 & 0.0 $\pm$ 0.0 \\
Fixed fusion ($\alpha_0{=}0.5$) & 9.26 $\pm$ 0.95 & 7.04 $\pm$ 0.65 & 6.96 $\pm$ 0.70 & 22.6 $\pm$ 6.4 & 2.8 $\pm$ 1.2 & 0.0 $\pm$ 0.0 \\
No reweight (rule; no shift) & 9.10 $\pm$ 0.92 & 7.00 $\pm$ 0.63 & 6.59 $\pm$ 0.67 & 28.0 $\pm$ 7.0 & 2.7 $\pm$ 1.0 & 0.0 $\pm$ 0.0 \\
With shift probe & 10.37 $\pm$ 1.71 & 8.13 $\pm$ 1.38 & 8.88 $\pm$ 1.59 & 22.6 $\pm$ 6.4 & 2.8 $\pm$ 1.2 & 24.8 $\pm$ 11.1 \\
\textbf{Proposed: rule fusion (no shift)} & \textbf{9.13 $\pm$ 0.90} & \textbf{7.04 $\pm$ 0.62} & \textbf{6.96 $\pm$ 0.68} & \textbf{22.6 $\pm$ 6.4} & \textbf{2.8 $\pm$ 1.2} & \textbf{0.0 $\pm$ 0.0} \\
\bottomrule
\end{tabular}}
\end{table*}

\begin{table*}[t]
\centering
\small
\caption{APC-inconsistent trips: mean $\pm$ std across 8 folds with at least three bad trips (std uses ddof=1). Rates are percentages; the shift rate is trip-level.}
\label{tab:main_bad_short}
\setlength{\tabcolsep}{2pt}
\resizebox{\textwidth}{!}{%
\begin{tabular}{lrrrrrr}
\toprule
Method & RMSE & MAE & Trip-end AE & Shadow infeas. & $e_{\mathrm{phys}}$ rate & Shift rate \\
\midrule
Perception-only (open-loop) & 37.67 $\pm$ 5.58 & 29.84 $\pm$ 5.79 & 35.10 $\pm$ 6.63 & 25.9 $\pm$ 11.3 & 0.0 $\pm$ 0.0 & 0.0 $\pm$ 0.0 \\
No fusion (phys-only) & 37.65 $\pm$ 4.58 & 29.70 $\pm$ 5.25 & 34.59 $\pm$ 6.18 & 27.1 $\pm$ 11.1 & 16.6 $\pm$ 9.1 & 0.0 $\pm$ 0.0 \\
Fixed fusion ($\alpha_0{=}0.5$) & 11.51 $\pm$ 1.97 & 8.63 $\pm$ 1.25 & 7.18 $\pm$ 1.39 & 27.1 $\pm$ 11.1 & 1.0 $\pm$ 0.7 & 0.0 $\pm$ 0.0 \\
No reweight (rule; no shift) & 10.55 $\pm$ 2.27 & 7.71 $\pm$ 1.47 & 6.31 $\pm$ 1.34 & 25.9 $\pm$ 11.3 & 1.0 $\pm$ 0.7 & 0.0 $\pm$ 0.0 \\
With shift probe & 11.85 $\pm$ 2.37 & 9.13 $\pm$ 1.47 & 7.41 $\pm$ 1.29 & 27.1 $\pm$ 11.1 & 1.0 $\pm$ 0.7 & 11.9 $\pm$ 9.8 \\
\textbf{Proposed: rule fusion (no shift)} & \textbf{10.57 $\pm$ 2.25} & \textbf{7.71 $\pm$ 1.46} & \textbf{6.40 $\pm$ 1.34} & \textbf{27.1 $\pm$ 11.1} & \textbf{1.0 $\pm$ 0.7} & \textbf{0.0 $\pm$ 0.0} \\
\bottomrule
\end{tabular}}
\end{table*}


\subsection{Coupling effects, trust allocation, and feasibility diagnostics}
The improvement among different tests is not caused by a linear stacking of modules. The fused state is written back and becomes the initial condition for the next stop, so the Perception--Physical--Fusion agents are coupled through shared state and residual signals. Removing key interaction edges leads to severe degradation (open-loop or phys-only), supporting that the gain comes from the stop-level recursion loop.

On the full test set, rule fusion is slightly better than fixed fusion ($\alpha_0=0.5$) in RMSE (9.13 vs.\ 9.26). The Physical Agent also provides an interpretable feasibility signal: compared with phys-only, the proposed system reduces the $e_{\mathrm{phys}}$ rate from $14.64\%$ to $2.81\%$, consistent with anchor-informed state correction reducing downstream over-alighting and denied-boarding events.

\subsection{Closed-loop calibration (residual-driven reweighting)}
We also test the training-level closed-loop calibration module by ablating residual-driven reweighting. Reweighting does not provide a consistent accuracy gain in RMSE/MAE in this dataset: the no-reweight variant achieves RMSE $9.10 \pm 0.92$ compared to $9.13 \pm 0.90$ for the proposed setting. Thus, reweighting is better interpreted as an auxiliary regularization module that reduces shadow infeasibility before physical projection, rather than as a primary driver of RMSE/MAE improvement. However, reweighting improves the drift diagnostic of the unconstrained shadow process: Shadow infeas.\ decreases from $27.97\% \pm 7.03\%$ (no reweight) to $22.57\% \pm 6.38\%$ (proposed). This indicates that residual feedback acts primarily as a stabilizer for drift-prone contexts rather than as a guaranteed accuracy-improving module under the present data and route conditions.

\subsection{Trajectory-level shift probe (macro-correction ablation)}
We evaluate the optional trajectory-level global shift probe as an ablation. Enabling this module increases overall error (RMSE $10.37 \pm 1.71$) and triggers on $24.76\% \pm 11.09\%$ of trips, suggesting over-correction in normal operating conditions. Consequently, the proposed framework disables the shift probe by default, and improving the activation gate is left for future work.
\vspace{-1.0em}
\subsection{Stress test on APC-inconsistent trips}
Table~\ref{tab:main_bad_short} reports performance on APC-inconsistent trips. Under this stress test, open-loop methods fail (perception-only RMSE $37.67 \pm 5.58$), while the proposed closed-loop remains stable (RMSE $10.57 \pm 2.25$). Rule fusion improves over fixed fusion (RMSE $10.57$ vs.\ $11.51$), consistent with dynamic trust allocation providing added robustness when evidence streams conflict. Because only 8 of the 15 folds contain at least three APC-inconsistent trips, these results should be interpreted as targeted robustness evidence for the inconsistent subset observed in this dataset, rather than a complete characterization of all sensor-failure regimes.
\vspace{-0.3em}
\subsection{Context features}
POI semantics yields no consistent improvement on the full test set under the current feature and model settings (RMSE $9.20 \pm 0.92$ vs.\ $9.13 \pm 0.90$), but provides a small gain on APC-inconsistent trips (RMSE $10.47 \pm 2.54$ vs.\ $10.57 \pm 2.25$), suggesting that stop semantics can be more useful under stressed conditions. Adding meteorological variables does not yield consistent gains under the present data and route conditions (RMSE $9.22 \pm 0.93$ overall; RMSE $10.49 \pm 2.53$ on APC-inconsistent trips). We also use the ABM layer as a qualitative plausibility audit in the mechanism case study (Fig.~\ref{fig:mechanism_cases}); fold-level ABM metrics are omitted due to space limits.

\subsection{Representative failure-mode case studies}
Representative held-out cases in Fig.~\ref{fig:mechanism_cases} illustrate how feasibility projection, anchor downweighting, and ABM auditing support the aggregate results.

\section{Conclusion and Future Directions}
\label{sec:conclusion}

This paper presents a closed-loop, state-centric, multi-agent framework for robust passenger load estimation from heterogeneous data streams. The method treats load inference as a stop-level state-recursion problem and integrates three core elements: (i) a segment-based pipeline that enforces physical feasibility during state estimation, (ii) adaptive trust allocation across multi-source streams, and (iii) residual-feedback learning to improve stability under crowding and noisy sensing. 

Experimental results confirm that open-loop accumulation of stop-level flows is highly drift-prone and can produce infeasible trajectories. Enforcing feasibility inside the recursion and writing back the fused state at each stop stabilizes the load evolution and reduces downstream infeasibility events. Feasibility projection alone is insufficient when streams disagree; the main benefit arises from coupling projection with anchoring and adaptive trust allocation within the closed loop.

Dynamic trust allocation provides additional robustness under stressed conditions, performing comparably to fixed fusion on the full test set but more reliably on APC-inconsistent trips. Residual-driven reweighting does not consistently improve RMSE/MAE, yet it reduces the infeasibility rate of the unconstrained shadow trajectory, indicating a stabilizing role in drift-prone contexts. Optional trajectory-level shift correction tends to over-correct under normal conditions and should remain disabled unless stricter activation criteria are defined. Context features such as POI semantics provide limited gains overall but show small improvements under stress.

Future work will extend the evaluation across multiple routes, time periods, and sensing configurations, especially where load channels degrade or Wi-Fi anchors become sparse. Stress testing will be strengthened with synthetic APC perturbations, reduced anchor availability, and complementary degradation indicators to better assess whether residual-driven reweighting improves stability under distribution shift. We will also explore recurrent and Transformer-based Perception Agents while preserving the same physical projection and trust-fusion structure. Finally, trust policies and macro-correction gates should be refined to improve adaptability while keeping the decision rules explicit and auditable.

	\bibliographystyle{IEEEtran}
	\bibliography{root}
\end{document}